\documentclass[conference]{IEEEtran}
\IEEEoverridecommandlockouts

\newcommand{\printCCTS}{\textit{CausalConceptTS }} 

\usepackage{algorithmic}
\usepackage[ruled,vlined]{algorithm2e}

\newcommand{\heading}[1]{\noindent\textbf{#1}}

\usepackage[normalem]{ulem} 
\newcommand{\stkout}[1]{\ifmmode\text{\sout{\ensuremath{#1}}}\else\sout{#1}\fi}

\newcommand{\deleted}[1]{}

\newcommand{\deletedfloat}[1]{}

\usepackage{enumitem}

\usepackage{comment}

\usepackage{graphicx}
\usepackage{amsmath}

\usepackage{amsmath, amssymb, amsfonts, amsthm}
\usepackage{graphicx}
\usepackage{amsfonts}       
\usepackage{nicefrac}       
\usepackage{microtype}      
\usepackage{mathtools}
\usepackage{comment}
\usepackage{pbox}
\usepackage{textcomp}
\usepackage[utf8]{inputenc}

\usepackage{nameref}
\usepackage{textalpha}
\usepackage{booktabs} 
\usepackage{comment}
\usepackage{subfig}

\usepackage{amssymb}
\usepackage{amsmath}
 \usepackage{url}

\usepackage{cite}
\usepackage{amsmath,amssymb,amsfonts}
\usepackage{algorithmic}
\usepackage{graphicx}
\usepackage{textcomp}
\usepackage{xcolor}
\def\BibTeX{{\rm B\kern-.05em{\sc i\kern-.025em b}\kern-.08em
    T\kern-.1667em\lower.7ex\hbox{E}\kern-.125emX}}
\begin{document}

\title{Explaining Time Series Classification Predictions via Causal Attributions\thanks{
© 2025 IEEE. Personal use of this material is permitted. Permission from IEEE must be obtained for all other uses, in any current or future media, including
reprinting/republishing this material for advertising or promotional purposes, creating new collective works, for resale or redistribution to servers or lists, or
reuse of any copyrighted component of this work in other works.}}

\author{\IEEEauthorblockN{1\textsuperscript{st} Juan Miguel Lopez Alcaraz}
\IEEEauthorblockA{\textit{AI4Health Division} \\
\textit{Carl von Ossietzky Universität Oldenburg}\\
Oldenburg, Germany \\
juan.lopez.alcaraz@uol.de}
\and
\IEEEauthorblockN{2\textsuperscript{nd} Nils Strodthoff}
\IEEEauthorblockA{\textit{AI4Health Division} \\
\textit{Carl von Ossietzky Universität Oldenburg}\\
Oldenburg, Germany \\
nils.strodthoff@uol.de}
}

\maketitle

\begin{abstract}
Despite the excelling performance of machine learning models, understanding their decisions remains a long-standing goal. Although commonly used attribution methods from explainable AI attempt to address this issue, they typically rely on associational rather than causal relationships. In this study, within the context of time series classification, we introduce a novel model-agnostic attribution method to assess the causal effect of concepts i.e., predefined segments within a time series, on classification outcomes. Our approach compares these causal attributions with closely related associational attributions, both theoretically and empirically. To estimate counterfactual outcomes, we use state-of-the-art diffusion models backed by state space models. We demonstrate the insights gained by our approach for a diverse set of qualitatively different time series classification tasks. Although causal and associational attributions might often share some similarities, in all cases they differ in important details, underscoring the risks associated with drawing causal conclusions from associational data alone. We believe that the proposed approach is also widely applicable in other domains to shed some light on the limits of associational attributions.
\end{abstract}

\begin{IEEEkeywords}
Classification, Machine Learning, Deep Learning, Time Series, Explainability, Causality.
\end{IEEEkeywords}

\section{Introduction}\label{sec: introduction}
Machine learning has advanced across diverse fields, driven by powerful hardware and large datasets. Time series data, prevalent in natural sciences, medicine, and life sciences \cite{Wang2023, esteva2019guide, miotto2018deep, shen2017deep, bepler2021learning}, enable modeling of temporal patterns, especially for classification tasks \cite{rajkomar2018scalable, Wang_2019}. Yet, complex models like deep learning often trade interpretability for performance, a critical concern in downstream applications \cite{Somani2021, roy2019deep}.

\heading{Need for Explainability}
Limited insight into model decisions hinders deploying deep learning models, especially in safety-critical areas. This challenge has driven the rise of explainable AI (XAI) \cite{lundberg2017unified,Montavon2018,covert2021explaining}. For time series classifiers, various XAI methods have been proposed \cite{crabbe2021explaining,raykar2023tsshap,zhao2023interpretation,ismail2020benchmarking}, but most rely on associations rather than causal effects. A clear theoretical and empirical understanding of how these attribution types differ remains missing.

\heading{Need for Causal Insights}
Counterfactual inference estimates the effect of an intervention by comparing observed outcomes with hypothetical alternatives. In medical applications, it helps assess treatment effects on patient outcomes \cite{gillies2018causality}. As noted in \cite{goyal2020explaining}, causal attributions are especially valuable with correlated features, where associational attributions may misrepresent the model's true decision basis.

\heading{Use-Cases for XAI} As outlined in \cite{wagner2023explaining}, XAI serves different purposes: user support, model auditing, and knowledge discovery. While user support requires user studies, auditing and discovery depend on faithful attributions that truly reflect model behavior. Relying on misleading associational attributions poses risks, especially in safety-critical domains. It is therefore essential to distinguish between causal and associational attributions when evaluating attribution methods.

\heading{Main Contributions}
We introduce Causal Concept Time Series Explainer (CausalConceptTS), a model-agnostic causal attribution method that improves interpretability in time series classification using predefined causal concepts as time series segments. Designed to be architecture-independent, it can be applied flexibly across models. Here, we demonstrate it using a diffusion-based imputer and structured state space classifiers. Our main contributions are:

\begin{itemize}
    \item[(1)] We formalize the distinction between causal and associational attributions for concept-based explanations in time series data.
    \item[(2)] We show how counterfactual outcomes, required for causal attributions, can be estimated using generative models such as diffusion-based imputers.
    \item[(3)] We provide a comparative analysis of causal and associational attributions across a diverse set of time series classification tasks, highlighting the limitations of purely associational explanations and the need for causal insights.
\end{itemize}

\begin{figure*}[!ht] 
\centering 
\includegraphics[width=\textwidth]{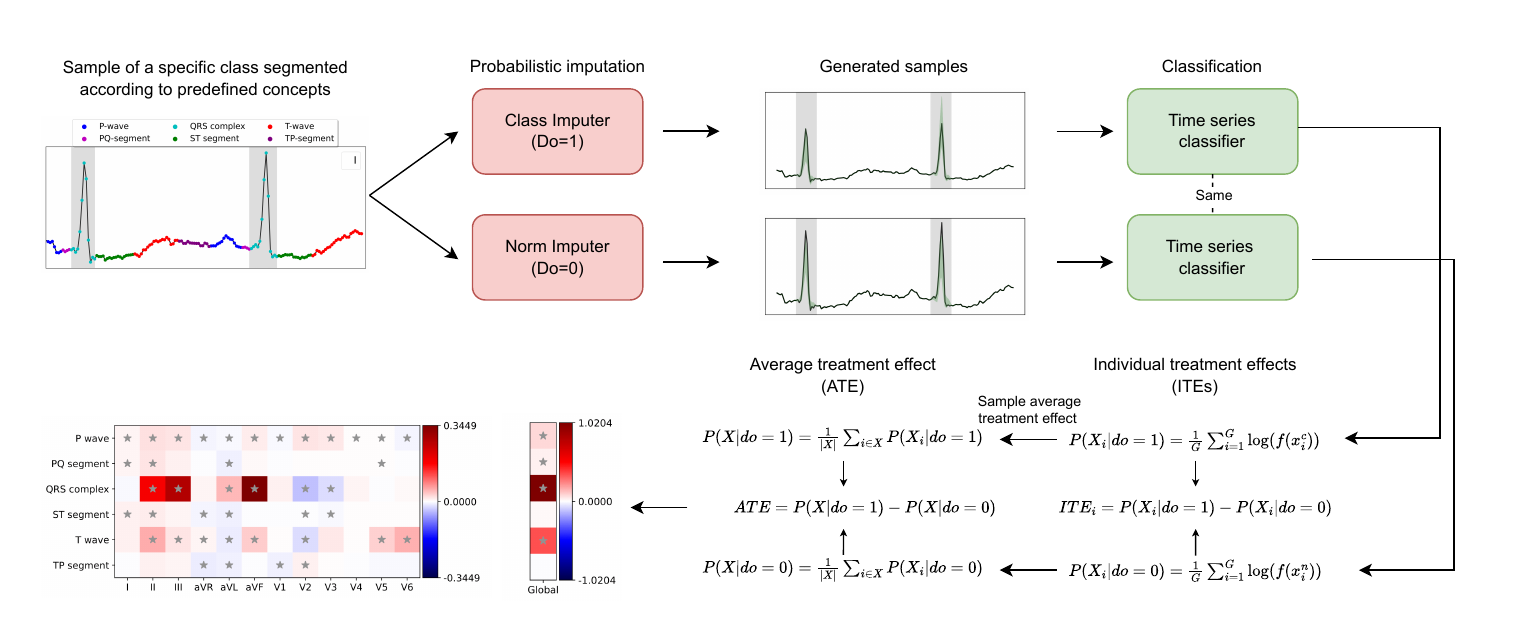} 
\caption{Schematic representation of the proposed \printCCTS approach. Starting from a sample belonging to a specific class, the time series is segmented into predefined concepts, either expert-defined (e.g., ECG segments) or inferred via clustering. For a selected concept, we generate counterfactual versions by imputing the corresponding segments using two different imputation models: one trained on samples from the original class, and another from a chosen baseline class (typically healthy controls). This process yields two sets of counterfactual samples, which are passed through a predefined classifier that we aim to investigate. The log difference between the mean output probabilities for the two sets yields an individual treatment effect (ITE),  a causal attribution quantifying the effect of the concept on the classifier's output. By averaging the ITEs across samples, we obtain the average treatment effect (ATE), which we visualize using both channel-agnostic and channel-specific causal attribution maps.} 
\vspace{1.2em}
\label{fig:abstract} 
\end{figure*}

\section{Background}\label{app: background}

\heading{Time Series Classification} 
Traditional time series classification encompasses distance-based \cite{rakthanmanon2013fast}, feature-based \cite{fulcher2017hctsa}, interval-based \cite{deng2013time}, shapelet-based \cite{hills2014classification}, and dictionary-based methods \cite{schafer2015boss}. Beyond these, deep learning approaches employ backbones such as Convolutional Neural Networks (CNNs) \cite{ismail2020inceptiontime}, Recurrent Neural Networks (RNNs) \cite{karim2017lstm}, self-attention mechanisms \cite{russwurm2020self}, and state space models \cite{Gu2021EfficientlyML}. This work focuses on the latter but is applicable to any classifier, including non-deep-learning models.

\heading{Deep Generative Models}
Deep learning has been applied to synthetic time series generation for tasks such as conditional generation \cite{Alcaraz:2023Synthetic}, class imbalance \cite{hssayeni2022imbalanced}, anomaly detection \cite{bashar2020tanogan}, imputation \cite{tashiro2021csdi, lopezalcaraz2023diffusionbased}, and explainability \cite{goyal2020explaining}. While early approaches used VAEs and GANs, diffusion models have recently become a strong alternative \cite{tashiro2021csdi, lopezalcaraz2023diffusionbased}. We adopt diffusion models to generate high-fidelity time series counterfactuals, though our method is compatible with other generative models and interventions, offering flexibility across applications.

\heading{Counterfactuals for Time Series Data}
Counterfactual analysis enables "what if" questions in time series classification by showing how input changes affect outcomes. Prior methods include random-sampling-based counterfactuals for multivariate treatment effects \cite{ates2021counterfactual}, which can produce discontinuities; instance-based interventions guided by class activation mappings \cite{delaney2021instance}, limited by network-derived regions; and motif-based interventions focused on recurring patterns \cite{li2022motif}. Latent-space counterfactual generation for univariate series has also been proposed \cite{wang2021learning}. To our knowledge, we are the first to estimate counterfactual time series inputs using high-fidelity diffusion models.

\heading{Attribution Methods for Time Series} 
Attribution methods help identify which parts of a time series most influence model decisions, improving trust in tasks like classification \cite{crabbe2021explaining} and forecasting \cite{ma2024tsfeatlime}. Reviews cover post-hoc methods, backpropagation, perturbation, approximation \cite{zhao2023interpretation} and ante-hoc methods \cite{rojat2021explainable}. Most attribution techniques focus on associational effects, linking predictions to correlations rather than causality. Our method instead infers causal effects in time series data. Related work by \cite{goyal2020explaining} used variational autoencoders for counterfactuals in image classification but relied on manually defined concepts. In contrast, we use data-driven, input-derived concepts for broader, more flexible analysis.

\section{CausalConceptTS: Causal Concept Time Series Explainer}\label{sec: CausalConceptTS}

In Figure~\ref{fig:abstract}, we provide a schematic overview of the proposed approach. In the following sections, we introduce the key concepts in detail.

\heading{Causal Data Generating Process} Building on the causal attribution framework for image data proposed by \cite{goyal2020explaining}, we adopt the causal data-generating process described by \cite{Schoelkopf2012}. For each sample, represented as a two-dimensional matrix $X \in \mathbb{R}^{l \times k}$ (with $l$ time steps and $k$ channels), we assume the existence of a class state $CS$ characterized by several binary indicator variables. The data generation process occurs in two stages:

\textbf{(1) Concept Assignment:} A semantic mask $M \in [1, \ldots, C]^{l \times k}$ assigns each token in the input sequence to one of $C$ predefined concepts, conditioned on a given class state $CS$. Concepts are therefore assumed to be non-overlapping.

\begin{figure}[!t] 
\centering 
\includegraphics[width=\linewidth]{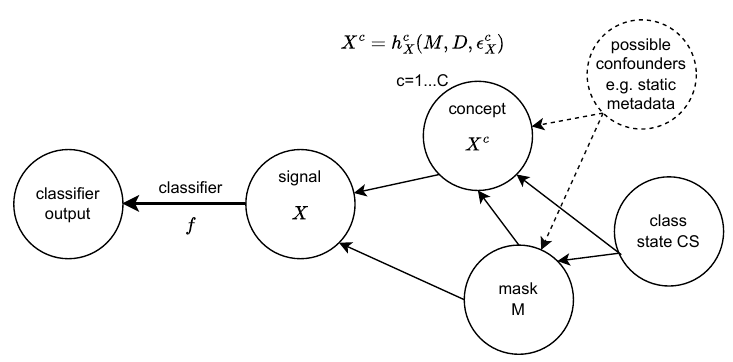} 
\caption{Causal graph underlying our approach. The data generating process starts from a class state $CS$, which influences a (concept) mask $M$. The combination of $CS$ and $M$ determines the specific numerical values $X^c_M$ for each concept $c$, leading to the input signal $X$. This input is passed through a predefined classifier $f$. We investigate the causal effect of $X^c_M$ on the classifier output by intervening on the class state $CS$. In our experiments, we assume that the class state $CS$ has no causal effect, and thus the concept mask $M$ is kept unchanged. We also omit potential confounders, such as static metadata that could influence $X^c_M$ or $M$, as indicated by the dashed lines.} 
\vspace{1.2em}
\label{fig:causalgraph} 
\end{figure}

\textbf{(2) Time Series Reconstruction:} The time series values corresponding to each concept $c \in {1, \ldots, C}$ are denoted as $X^c_M = {X_{ab} \mid M_{ab} = c}$. The full input sequence $X$ can then be reconstructed as $X = X(X^1_M, \ldots, X^C_M)$, where $X_{ab} = X^c_M[k]$ such that $c = M_{ab}$ and $k = |{(i, j) \mid M_{ij} = c \text{ and } (i,j) \preceq (a,b)}|$, with $\preceq$ denoting lexicographic ordering. We assume that each $X^c_M$ is generated through a structural causal model $h_X^c$, i.e., $X^c = h_X^c(M, CS, \epsilon_X^c)$, where $M$ is the semantic mask, $CS$ is the class state, and $\epsilon_X^c$ represents noise. Our goal is to study a predefined (binary) classifier $f$ that maps the input sequence $X$ to an output probability $f(X)$. A visualization of the causal graph underlying this process is shown in Figure~\ref{fig:causalgraph}.

\heading{Individual and Average Treatment Effects} We aim to investigate the causal effect of the class state \( CS \) on the classifier \( f \) by intervening on \( CS \). For simplicity, we assume that the underlying semantic mask \( M \) remains unchanged during this intervention for a given sample. Specifically, we intervene by setting the class state to a value \( CS^* \). The baseline class state \( CS^0 \) represents the reference (in a medical context, this could correspond to healthy control samples). The \textit{individual treatment effect (ITE)} for sample \( X \) of concept \( C \in [1, \ldots, C] \) on the classifier \( f \) is defined using do-operators as in \cite{pmlr-v70-shalit17a}:

\begin{align}
\begin{split}
\text{ITE}&(X,f,c,CS^*,CS^0) = \\
&\log_2 E_{h_X^c} f\left(X\left(X^{c\complement}_M,\ (X^c_M \mid \text{do}(CS=CS^*))\right)\right) \\
&-\  \log_2 E_{h_X^c} f\left(X\left(X^{c\complement}_M,\ (X^c_M \mid \text{do}(CS=CS^0))\right)\right)
\end{split}
\label{eq:ite}
\end{align}

where \( X^{c\complement}_M \) denotes the set of all \( X^i_M \) for \( i \neq c \). The expectation in Eq.~(\ref{eq:ite}) is taken over the data-generating process \( h_X^c \). We adopt logarithmic differences to compare output probabilities, as discussed in \cite{Bluecher:2021PredDiff} in the context of associational attributions. To obtain the \textit{average treatment effect (ATE)}, we average the ITE over the distribution of samples with label \( CS^* \):
\begin{align}
\begin{split}
\text{ATE}&(f,c,CS^*,CS^0) = \\ 
&E_{X \sim \mathcal{X}(CS^*)} \text{ITE}(X,f,c,CS^*,CS^0)
\end{split}
\label{eq:ate}
\end{align}
where \( \mathcal{X}(CS^*) \) refers to the data distribution of samples with class state \( CS^* \).

\heading{Individual Associational Effect} The individual treatment effect is structurally similar to the \textit{PredDiff} attribution method \cite{Bluecher:2021PredDiff}, which can be seen as a special case of the Shapley value where only a single coalition (the complement of the feature set \( X^c_M \)) contributes. In analogy to Eq.~(\ref{eq:ite}), we define the \textit{individual associational attribution (IAA)} as:
\begin{align}
\begin{split}
\text{IAA}&(X,f,c,CS^*,CS^0) =\\
&\log_2 f(X) 
- \log_2 E_{x^c_M \sim k_X^c} f\left(X\left(X^{c\complement}_M, x^c_M\right)\right)
\end{split}
\label{eq:preddiff}
\end{align}
where the expectation refers to the conditional distribution \( k_X^c \equiv p(X^c_M \mid X^{c\complement}_M) \). The IAA coincides with the PredDiff attribution for \( X^c_M \).

\heading{Relation Between Causal and Associational Attributions} To explore the differences and similarities between causal and associational attributions, we compare Eq.~(\ref{eq:ite}) and Eq.~(\ref{eq:preddiff}). The first term in Eq.~(\ref{eq:ite}) represents the observed outcome, and we expect that $\log_2 E_{h_X^c}f(X({X^{c\complement}_M},(X^c_M|\text{do}(CS=CS^*))))\approx$ $f(X)$ if $CS^*$ coincides with the true label of the sample $X$. The second term refers to the counterfactual outcome. The primary difference between causal ITE (Eq.~(\ref{eq:ite})) and associational attribution (Eq.~(\ref{eq:preddiff})) lies in the use of a class-conditional imputer for the causal ITE

\begin{equation}
\begin{split}
E_{h_X^c}& f\left(X\left(X^{c\complement}_M,\ (X^c_M \mid \text{do}(CS=CS^0))\right)\right) 
\approx\ \\
&\int f\left(X\left(X^{c\complement}_M, x^c_M\right)\right) 
p(x^c_M \mid X^{c\complement}_M, CS^0)\, dx^c_M
\end{split}
\label{eq:conditional}
\end{equation}

compared to the class-unconditional imputer used in the associational IAA
\begin{equation}
\begin{split}
E_{x^c_M \sim k_X^c}& f\left(X\left(X^{c\complement}_M, x^c_M\right)\right) 
= \\
&\int f\left(X\left(X^{c\complement}_M, x^c_M\right)\right) 
p(x^c_M \mid X^{c\complement}_M)\, dx^c_M
\end{split}
\label{eq:unconditional}
\end{equation}

This distinction allows us to empirically compare causal and associational attributions at the level of individual samples. The relation in Eq.~(\ref{eq:conditional}) is only approximate as it captures the dependence of the generative distribution on \( CS^0 \) but ignores other potential confounders, such as static metadata. In the supplementary material, we demonstrate for a simple example involving a single concept that associational attributions can lead to unnatural outcomes, where the attribution changes sign, while the causal attribution shows no such behavior.

\heading{Generative Model Architecture} We specify the generative model used to sample from the unconditional distribution \( h_X^c \) or the conditional distribution \( k_X^c \), as approximated in Eq.~\ref{eq:unconditional} and Eq.~\ref{eq:conditional}. We employ the structured state-space diffusion (\textit{SSSD}) model for time series imputation \cite{lopezalcaraz2023diffusionbased}, an extension of \textit{DiffWave} \cite{DBLP:conf/iclr/KongPHZC21} that replaces bidirectional dilated convolutions with two S4 layers to better capture long-term dependencies. Combined with a modified diffusion procedure that applies noise only to concepts requiring imputation, this model achieves state-of-the-art imputation performance. Class-conditional models are trained by subsampling the training set by label, approximating \( h_X^c \) and \( k_X^c \) with some sampling error. Prediction intervals are estimated from multiple imputations \cite{lopezalcaraz2023diffusionbased}; while conformal prediction \cite{angelopoulos2021gentle} provides valid intervals in-distribution, guarantees do not hold for counterfactuals due to distributional shift.

\heading{Generative Model Details}  The imputer in \printCCTS uses 36 residual layers with 256 residual and skip channels, maintaining other hyperparameters from \cite{lopezalcaraz2023diffusionbased}. We optimize mean squared error (MSE) with Adam over 200 diffusion steps using a linear noise schedule. Expectation values in Eq.~\ref{eq:conditional} and Eq.~\ref{eq:unconditional} are approximated through sampling, converging after about 15 samples; for robustness, we generate 40 samples per real instance. Training details and additional details on the computational complexity can be found in the supplementary material. 

\heading{Channel-Specific Attributions} To assess channel-specific attributions, we avoid conditioning on inputs from other channels captured at the same time to prevent bias from correlated channels \cite{Bluecher:2021PredDiff}. We use an imputer trained in a blackout-missing setting and substitute non-targeted channels with their original values, where blackout-missing scenario is defined as a single segment across all channels, i.e., the respective segment is assumed to be missing across all channels.

\heading{Classifier Model Architecture} Following successful applications in physiological time series \cite{10.1093/ehjdh/ztae039,alcaraz2024mds} where it compared preferably compared to modern CNN-based classifiers, we employ structured state-space models (S4) with four layers as classifiers \cite{Gu2021EfficientlyML}. Optimization uses Adam with a constant learning rate, weight decay of 0.001, batch size 64, and 20 training epochs, with binary cross-entropy as the loss. Model selection is based on the best validation set score, typically converging early. We use S4 as the classifier backbone due to its efficiency and ability to model temporal dependencies across multiple scales. We stress again that the proposed framework works irrespective of the chosen classifier architecture, also for non-Deep Learning models.

\heading{Performance Metric}
We use AUROC as the primary metric because it effectively measures discriminative ability without predefined thresholds and handles imbalanced data well. Recent research \cite{mcdermott2024closer} supports AUROC's superiority over metrics like AUPRC in such scenarios. It is important to note that our attributions are based on class probabilities, not directly on AUROC, as the metric is primarily used to demonstrate the classifier's predictive capabilities. Test set results are reported with 95\% confidence intervals from 1,000 bootstrapped iterations. For additional details on the classifier model, we refer to the supplementary material.

\heading{Concept Discovery and Validation} 
When predefined concepts are unavailable, we perform discovery via k-means clustering on raw time series using squared Euclidean distance, selecting the number of clusters via simple heuristics. Here, k-means is applied to sliding windows of length ``sample length''. It is worth stressing again that the focus of this paper lies on causal attribution rather than concept discovery. Concepts discovered via clustering only demonstrate the applicability of the method in the absence of expert-defined concepts. This is also the reason why we refrain from using more complicated sliding measures, which could improve the handling of time series with global misalignment. We validate the semantic meaning of the discovered concepts by training an XGBoost classifier on six statistical features (minimum, maximum, mean, standard deviation, median, number of time steps) extracted per cluster. Concept discovery is validated through classification performance rather than unsupervised clustering metrics, with high performance indicating effective concept separation and practical utility for downstream tasks.

\heading{Uncertainty Quantification in ATEs} Approximating expectations via finite samples enables both point estimates and uncertainty quantification for ITEs/IAAs and ATEs. We perform 1,000 bootstrap iterations sampling with replacement from the test set to compute 95\% prediction intervals for ATEs. A causal effect is considered statistically significant if the interval excludes zero.

\section{Experiments}\label{sec: experiments}

\heading{Structure of This Section} 
We conduct experiments across diverse time series classification tasks, reporting results for three meteorological and physiological datasets. Dataset details and preprocessing steps are provided in the supplementary material. We would like to emphasize that only one (drought prediction) out of the three considered datasets lacks expert concepts and therefore relies on concepts identified via clustering. Results are presented through figures showing associational and causal attributions: the right side displays global causal effects across all channels, while the left side shows channel-specific treatment effects per concept. Statistically significant effects, where the 95\% prediction interval excludes 0, are marked with a star. Exemplary time series plots with superimposed concept assignments illustrate the concepts. Details on the computational complexity of our approach are provided in Section~\ref{app:computation} of the supplementary material, while information on the datasets and additional model hyperparameters can be found in Sections~\ref{app: datasets} and~\ref{app: models}, respectively. To support further research, we provide the source code and supplementary material in a public repository: \url{https://github.com/AI4HealthUOL/CausalConceptTS}.

\heading{Goal of This Study} 
This work focuses on comparing associational and causal attributions. Specifically, we compare with associational PredDiff attributions, the analogues of our proposed causal attributions. We do not compare our method with structurally dissimilar associational attribution approaches, as such comparisons would detract from the primary focus of this study. For a broader comparison of associational methods, see \cite{bluecher2024decoupling}. Our model provides causal attributions for a given classifier. Assuming the classifier captures true relationships, we directly compare its outputs to ground-truth causal effects established in the literature, aiming for alignment with the underlying data. We emphasize that our primary objective is faithfulness of attributions, assessed via consistency with established effects rather than human-centered evaluation through user studies. However, faithful explanations can also support human decision-making by highlighting clinically relevant patterns.

\begin{figure}[!ht]
    \centering
    \includegraphics[height=0.25\textheight,width=\linewidth]{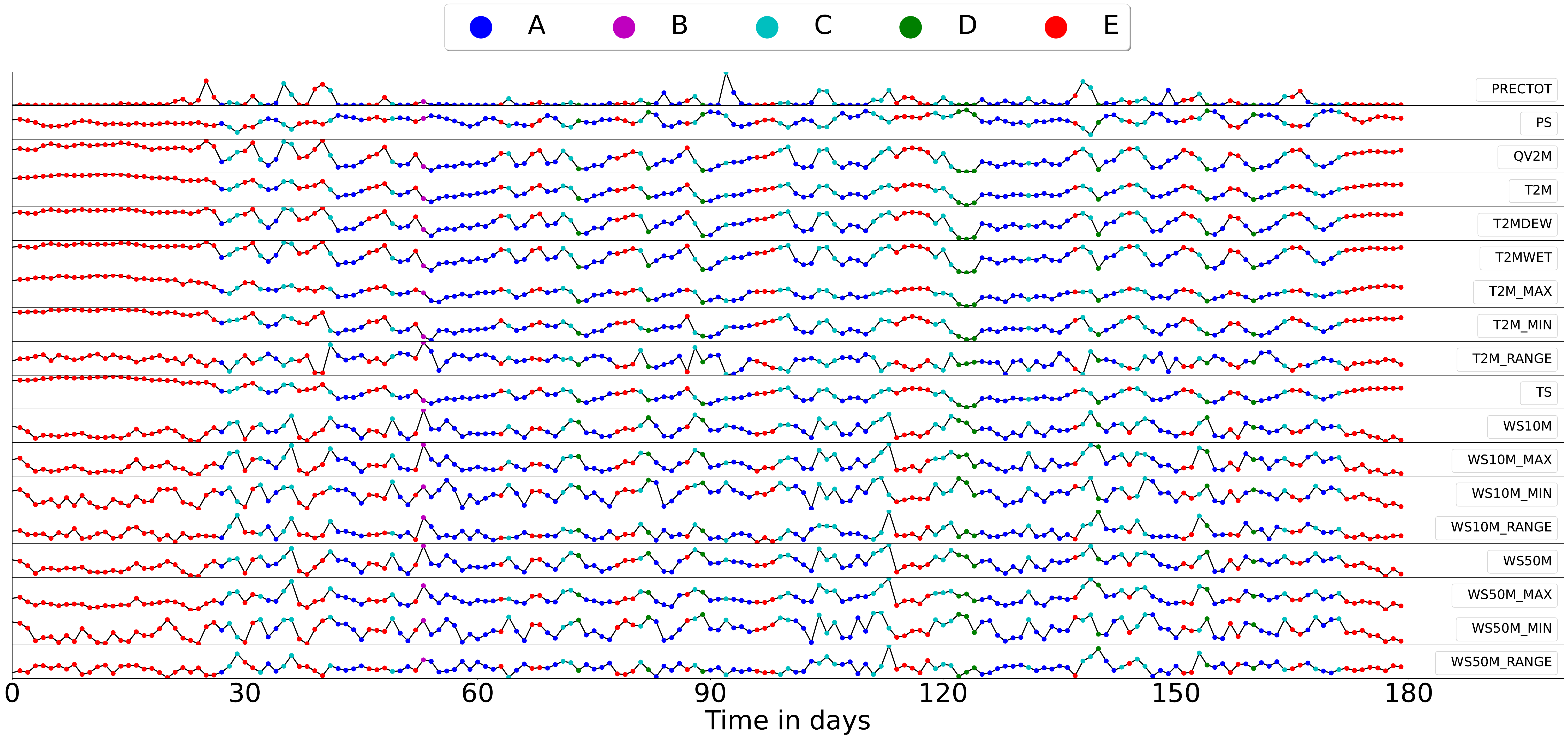}
    \caption{Schematic representation of identified concepts in the drought dataset for a sample encompassing all channels. Concept assignments (A-E) were derived using k-means clustering.}
    \vspace{1.2em}
    \label{fig:concept_drought_s}
\end{figure}
\vspace{1.2em}

\heading{Drought Prediction} As a first task, we use the drought dataset \cite{Minixhofer_2021} from the U.S. Drought Monitor to classify whether the upcoming week will experience drought, based on six months of daily meteorological data with 18 features (e.g., minimum and maximum precipitation, pressure, humidity, temperature, and wind speed at different meters, see supplementary material for details). In the absence of expert-defined concepts, we identify five concepts (A-E) via k-means clustering, achieving an AUROC of 0.7447 (95\% PI: 0.7406-0.7483) during concept validation. The S4 classifier reaches an AUROC of 0.8941 (95\% PI: 0.8919-0.8962). Concept assignments for a sample are shown in Fig.~\ref{fig:concept_drought_s}.

\begin{figure}[!ht]
    \centering
    \begin{minipage}{\linewidth}
        \centering
        \includegraphics[width=\linewidth]{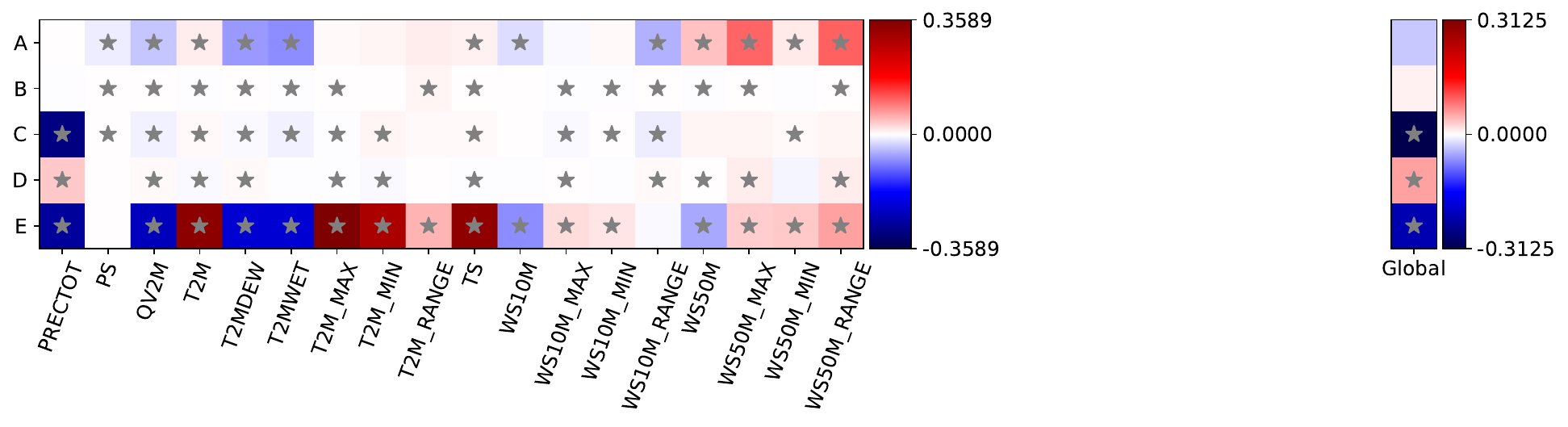}
        \caption*{(A) Associational attributions}
        \vspace{1.2em}
    \end{minipage}
    
    \begin{minipage}{\linewidth}
        \centering
        \includegraphics[width=\linewidth]{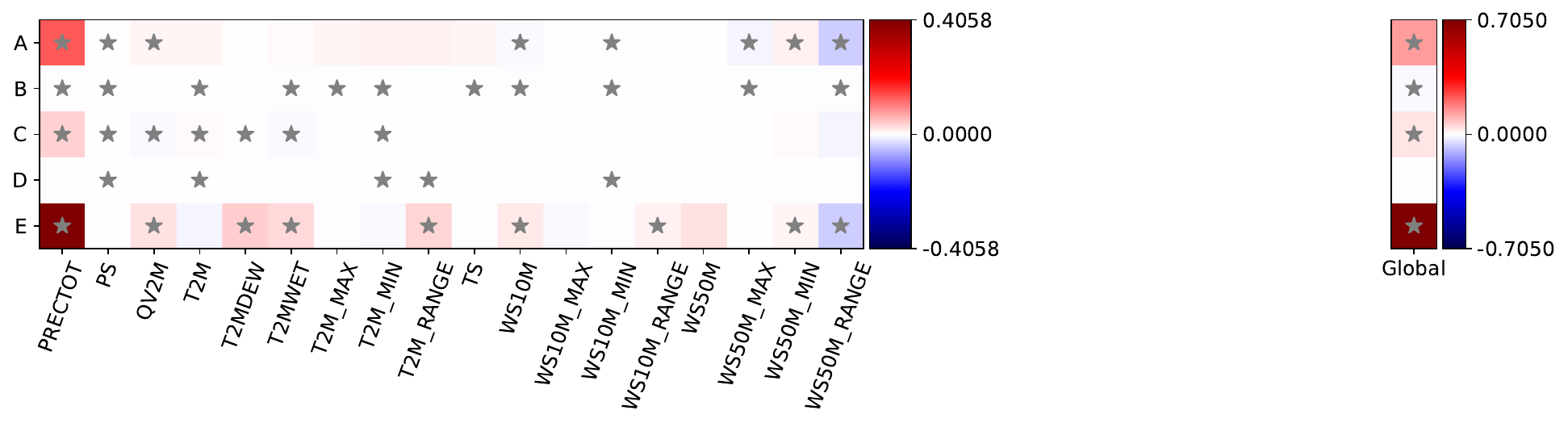}
        \caption*{(B) Causal attributions}
        \vspace{1.2em}
    \end{minipage}
    
    \caption{Illustration of the (A) associational and (B) causal attributions on the drought dataset. In both panels, rows correspond to time series channels and columns to concepts. Color intensity represents the direction and magnitude of attribution: dark red indicates a strong positive effect, dark blue a strong negative effect, and white indicates no effect. Stars (*) mark statistically significant attributions (95\% prediction interval excluding zero).}
    \vspace{1.2em}
    \label{fig:ate_drought}
\end{figure}

Fig.~\ref{fig:ate_drought} presents (A) associational and (B) causal attributions for the drought prediction task. Both channel-wise maps identify diverse variables with significant effects but sometimes differ in effect direction. Notably, precipitation shows the highest positive effect causally but a negative associational effect. This aligns with extensive studies confirming precipitation's positive impact on drought prediction \cite{cancelliere2007drought, anshuka2019drought}. Similarly, in concept E, variables at 2 meters such as humidity, dew/frost point, and wet bulb temperature have positive effects recognized by causal but not associational attributions \cite{behrangi2015utilizing}. For concept A, wind speed metrics at 50 meters (minimum, maximum, range) positively influence drought, as correctly attributed by causal methods \cite{vstvepanek2018drought}, unlike associational attributions.

\begin{figure}[!ht]
    \centering
    \includegraphics[height=0.25\textheight,width=\linewidth]{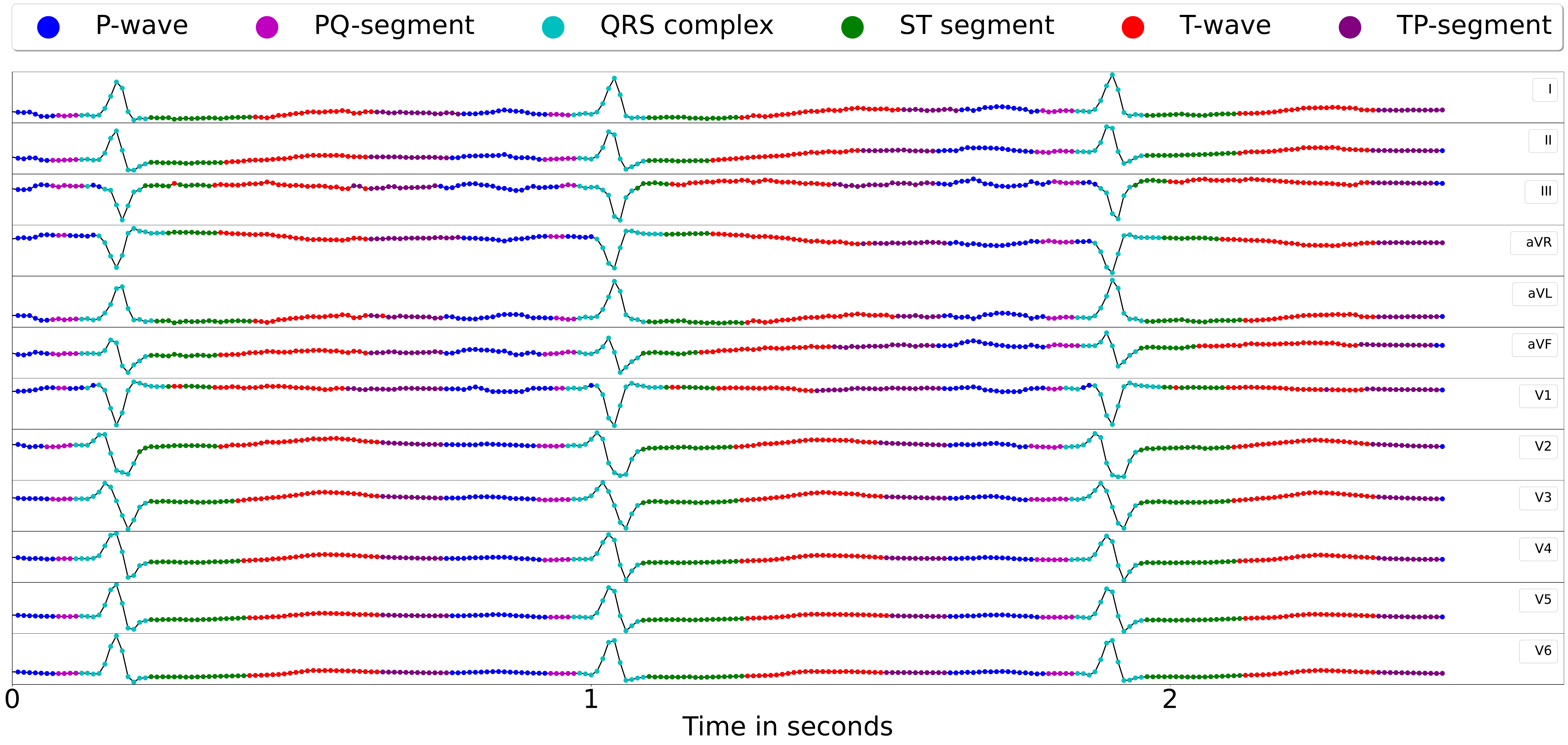}
    \caption{Schematic representation of identified concepts in the ECG dataset for a sample encompassing all channels. Concept assignments (P-wave, PQ segment, QRS complex, ST segment, T wave, and TP segment) were collected from the segmentation maps provided by \cite{wagner2023explaining}.}
    \vspace{2em}
    \label{fig:concept_ptbxl_s}
\end{figure}

\heading{ECG Classification}  For the second dataset, we use the PTB-XL dataset \cite{Wagner:2020PTBXL, Goldberger2020:physionet}, containing clinical 12-lead ECG data. Although PTB-XL includes multi-label hierarchical ECG annotations, we focus on binary classification of inferior myocardial infarction (IMI) versus healthy controls (NORM+SR). Using 248 time steps and channel-specific ECG segmentations from \cite{wagner2023explaining}, we consider six concepts: P-wave, PQ-segment, QRS complex, ST-segment, T-wave, and TP-segment, which achieve an AUROC of 0.9287 (95\% PI 0.913-0.9435) during concept validation. The classifier attains an AUROC of 0.9722 (95\% PI 0.9621-0.9797). Fig.~\ref{fig:concept_ptbxl_s} visualizes these concepts for a myocardial infarction sample.

\begin{figure}[!ht]
    \centering
    \begin{minipage}{\linewidth}
        \centering
        \includegraphics[width=\linewidth]{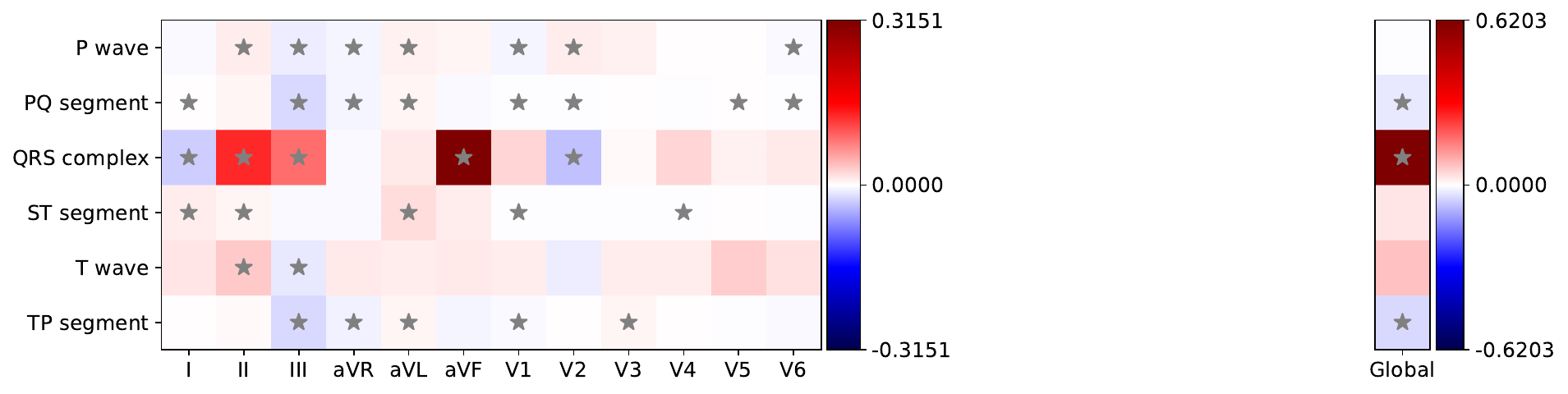}
        \caption*{(A) Associational attributions}
        \vspace{1.2em}
    \end{minipage}
    
    \begin{minipage}{\linewidth}
        \centering
        \includegraphics[width=\linewidth]{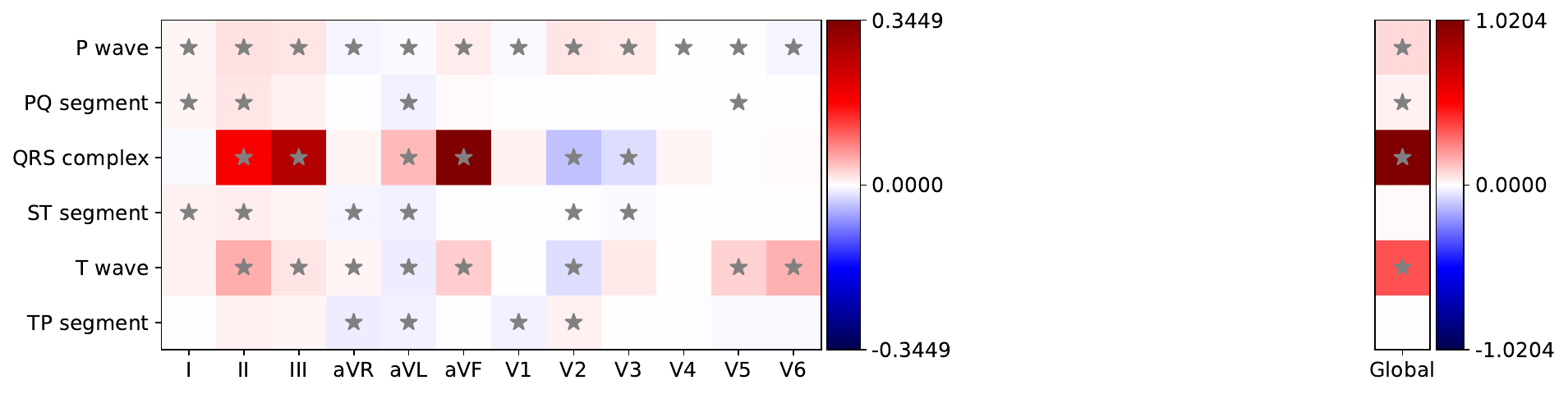}
        \caption*{(B) Causal attributions}
        \vspace{1.2em}
    \end{minipage}
    
    \caption{Illustration of the (A) associational and (B) causal attributions on the PTB-XL dataset. For a description of the figure layout and notation, see the caption of Fig.~\ref{fig:ate_drought}.}
    \vspace{2em}
    \label{fig:ate_ptbxl}
\end{figure}

Fig.~\ref{fig:ate_ptbxl} shows associational and causal attributions for the ECG classification task. Both maps highlight positive effects of the QRS complex in leads II, III, and aVF, consistent with pathological Q-waves \cite{thygesen2018fourth}. Associational attributions show a negative significant effect on the T-wave in lead III, whereas causal attributions show a positive effect, aligning better with literature that associates high T-waves with positive outcomes \cite{dressler1947high}. Similarly, causal attributions recognize positive effects for the P-wave in leads I, II, and III \cite{grossman1969serial}, while associational attributions show positive effects only in lead II and a negative effect in lead III.

\begin{figure}[!ht]
\centering
\includegraphics[width=\linewidth]{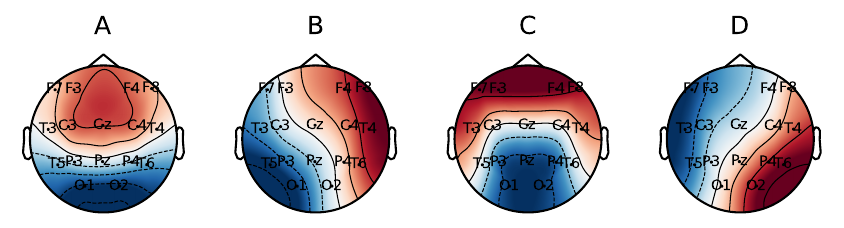}
\caption{Spatial distribution of brain activity patterns during different states of brain processing. Dark red indicates increased activity, while dark blue signifies decreased activity.}
\vspace{1.2em}
\label{fig: topographic map4}
\end{figure}

\begin{figure}[!ht]
    \centering
    \includegraphics[height=0.25\textheight,width=\linewidth]{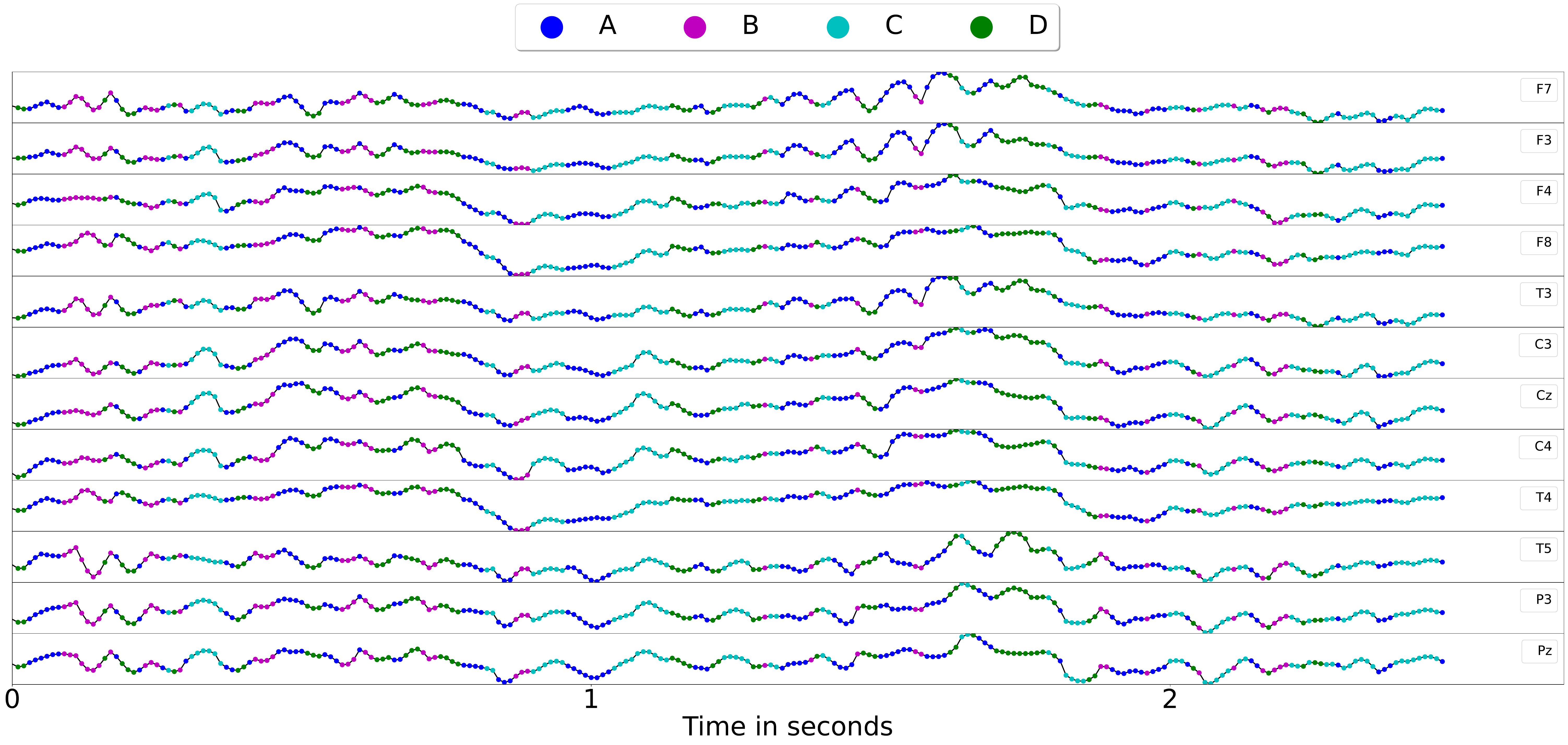}
    \caption{Schematic representation of identified concepts in the schizophrenia dataset for a sample encompassing all channels. Concept assignments (A-E) were derived using k-means clustering.}
    \vspace{1.2em}
    \label{fig:concept_schizophrenia_s}
\end{figure}

\heading{EEG Classification}  As the third dataset, we analyze the schizophrenia EEG dataset \cite{borisov2005analysis}, comprising 16 channels with 248 time steps each, from paranoid schizophrenia patients and healthy controls. Concepts are extracted via EEG microstate analysis \cite{pascual1995segmentation} using open-source tools \cite{githubGitHubFredericvWeegmicrostates, GramfortEtAl2013a}, capturing transient brain states linked to neural dynamics and cognition. Four concepts (A-D) yield a concept validation AUROC of 0.8249 (95\% PI 0.7682–0.8793). Figure~\ref{fig: topographic map4} shows a topographic map of brain activity per microstate for comparison with literature. The S4 model achieves classification AUROC of 0.9671 (95\% PI 0.9432–0.9849). Fig.~\ref{fig:concept_schizophrenia_s} visualizes concept assignments for a schizophrenia sample.

\begin{figure}[!ht]
    \centering
    \begin{minipage}{\linewidth}
        \centering
        \includegraphics[width=\linewidth]{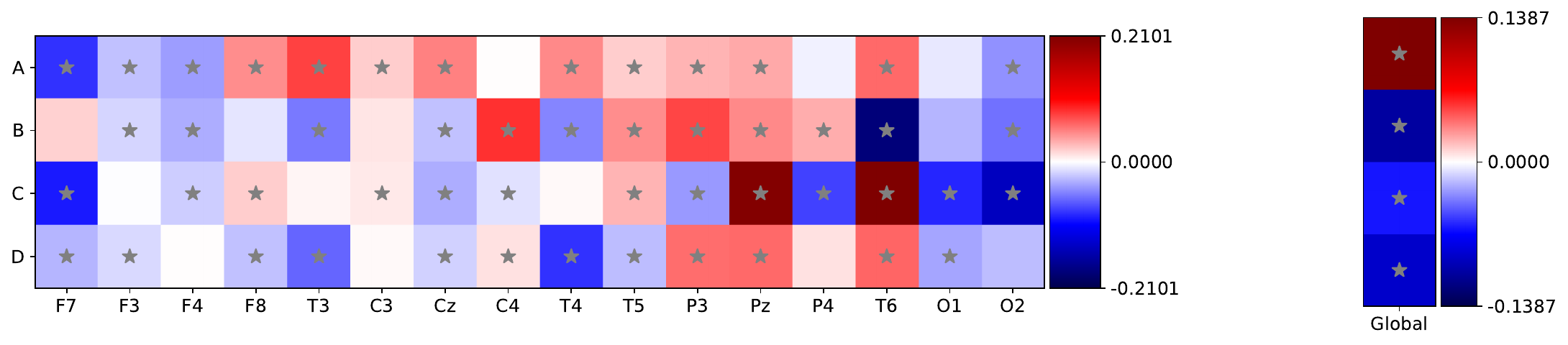}
        \caption*{(A) Associational attributions}
        \vspace{1.2em}
    \end{minipage}
    
    \begin{minipage}{\linewidth}
        \centering
        \includegraphics[width=\linewidth]{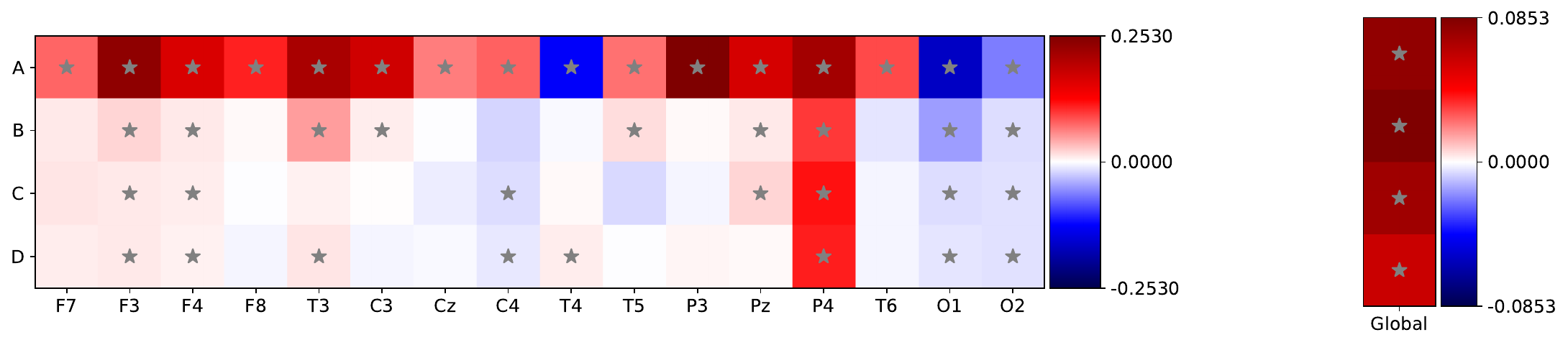}
        \caption*{(B) Causal attributions}
        \vspace{1.2em}
    \end{minipage}
    
    \caption{Illustration of the (A) associational and (B) causal attributions on the schizophrenia dataset. For a description of the figure layout and notation, see the caption of Fig.~\ref{fig:ate_drought}.}
    \vspace{2em}
    \label{fig:ate_eeg}
\end{figure}

Fig.~\ref{fig:ate_eeg} shows associational and causal attributions for EEG classification. Literature identifies concept B as significantly different between schizophrenia patients and controls across multiple studies considering duration \cite{KIKUCHI2007163, koenig1999deviant, NISHIDA20131106} and occurrence \cite{koenig1999deviant, NISHIDA20131106}. Concepts A and C are also highlighted for features like occurrence and coverage \cite{2022_resting}, and concept D for increased mean duration \cite{10.3389/fpsyt.2021.761203}. Globally, causal attributions align better with expert knowledge than associational ones. Notably, from a channel-wise perspective, this work is the first to investigate single-lead microstate effects in schizophrenia EEG classification; unlike previous datasets, associational attributions here appear inconsistent across channels.

\heading{Toy Example of Causal vs. Associational Attribution}
To illustrate the difference between causal and associational attributions, we considered a simple single-concept toy example involving a sine-wave signal with class-dependent amplitude. The full setup and derivations are provided in Section~\ref{app:toy} of the supplementary material.

\section{Discussion and Conclusion}\label{sec: discussion}

\heading{Limitations} 
At this stage, \printCCTS has several limitations. First, it does not account for intervening on the concept mask $M$, i.e., varying concept length per class, but relies on a predefined $M$ from the original sample. This may be problematic for pathologies like left bundle branch block in ECG, characterized by a wide QRS complex. Combining adjacent concepts instead of individual ones could mitigate this. Second, the generative model is trained only on real samples, assuming it generalizes to unseen classes when conditioned on concepts from other classes, an inherent challenge in causal estimation. Intervening on concepts from a different class pushes the model slightly outside its scope, blending original and intervened characteristics. Third, the method focuses solely on causal effects of concepts, without adjusting for confounding factors such as static patient metadata; their influence could be addressed by conditioning the generative model on these variables.

\heading{Future Work}
We use diffusion and state space models for imputation due to their strong performance and focus primarily on the framework rather than improving reconstruction quality. Future work should evaluate diverse architectures to assess their effect on causal and associational attributions, as explored in computer vision \cite{bluecher2024decoupling}. Investigating concept interactions across time and channels such as how earlier concepts influence later ones, via methods like Granger causality is also important. Analyzing channel correlations and interaction effects from both causal and associational perspectives (e.g., \cite{Bluecher:2021PredDiff}) is critical for understanding time series relationships. Another key direction is developing synthetic datasets with known causal structures, addressing limitations of relying on expert-assumed ground truth. Simiarly, while our approach targets binary classification, it can be naturally extended to multiclass problems \cite{goyal2020explaining}. 

\heading{Conclusion} The paper proposes a framework to assess the causal effect of the class-specific manifestation of predefined concepts within a time series on a given fixed time series classifier. Its key component is a high-fidelity diffusion model, which is used to infer counterfactual manifestations of concepts under consideration. This allows us to compute individual and average treatment effects. Furthermore, we demonstrate that the main difference between such causal attributions and purely associational, perturbation-based attributions lies in the use of a class-conditional as opposed to an unconditional imputation model. These insights allow for a direct comparison of causal and associational attributions. The differences between causal and associational attributions hint at the danger of drawing misleading conclusions from associational attributions. We showcase our approach for a diverse set of three time series classification tasks and find a good alignment of the identified causal effects with expert knowledge.

\bibliographystyle{IEEEtran}
\bibliography{mybibfile}

@article{vstvepanek2018drought,
  title={Drought prediction system for Central Europe and its validation},
  author={{\v{S}}t{\v{e}}p{\'a}nek, Petr and Trnka, Miroslav and Chuchma, Filip and Zahradn{\'\i}{\v{c}}ek, Pavel and Skal{\'a}k, Petr and Farda, Ale{\v{s}} and Fiala, Rostislav and Hlavinka, Petr and Balek, Jan and Semer{\'a}dov{\'a}, Daniela and others},
  journal={Geosciences},
  volume={8},
  number={4},
  pages={104},
  year={2018},
  publisher={MDPI}
}

@article{behrangi2015utilizing,
  title={Utilizing humidity and temperature data to advance monitoring and prediction of meteorological drought},
  author={Behrangi, Ali and Loikith, Paul C and Fetzer, Eric J and Nguyen, Hai M and Granger, Stephanie L},
  journal={Climate},
  volume={3},
  number={4},
  pages={999--1017},
  year={2015},
  publisher={MDPI}
}

@article{Wang2023,
  doi = {10.1038/s41586-023-06221-2},
  year = {2023},
  month = aug,
  publisher = {Springer Science and Business Media {LLC}},
  volume = {620},
  number = {7972},
  pages = {47--60},
  author = {Hanchen Wang and Tianfan Fu and Yuanqi Du and Wenhao Gao and Kexin Huang and Ziming Liu and Payal Chandak and Shengchao Liu and Peter Van Katwyk and Andreea Deac and Anima Anandkumar and Karianne Bergen and Carla P. Gomes and Shirley Ho and Pushmeet Kohli and Joan Lasenby and Jure Leskovec and Tie-Yan Liu and Arjun Manrai and Debora Marks and Bharath Ramsundar and Le Song and Jimeng Sun and Jian Tang and Petar Veli{\v{c}}kovi{\'{c}} and Max Welling and Linfeng Zhang and Connor W. Coley and Yoshua Bengio and Marinka Zitnik},
  title = {Scientific discovery in the age of artificial intelligence},
  journal = {Nature}
}

@article{roy2019deep,
  title={Deep learning-based electroencephalography analysis: a systematic review},
  author={Roy, Yannick and Banville, Hubert and Albuquerque, Isabela and Gramfort, Alexandre and Falk, Tiago H and Faubert, Jocelyn},
  journal={Journal of neural engineering},
  volume={16},
  number={5},
  pages={051001},
  year={2019},
  publisher={IOP Publishing}
}

@article{Somani2021,
  doi = {10.1093/europace/euaa377},
  year = {2021},
  month = feb,
  publisher = {Oxford University Press ({OUP})},
  volume = {23},
  number = {8},
  pages = {1179--1191},
  author = {Sulaiman Somani and Adam J Russak and Felix Richter and Shan Zhao and Akhil Vaid and Fayzan Chaudhry and Jessica K De Freitas and Nidhi Naik and Riccardo Miotto and Girish N Nadkarni and Jagat Narula and Edgar Argulian and Benjamin S Glicksberg},
  title = {Deep learning and the electrocardiogram: review of the current state-of-the-art},
  journal = {{EP} Europace}
}

@inproceedings{
mcdermott2024closer,
title={A Closer Look at {AUROC} and {AUPRC} under Class Imbalance},
author={Matthew B.A. McDermott and Haoran Zhang and Lasse Hyldig Hansen and Giovanni Angelotti and Jack Gallifant},
booktitle={The Thirty-eighth Annual Conference on Neural Information Processing Systems},
year={2024},
url={https://openreview.net/forum?id=S3HvA808gk}
}

@article{
lopezalcaraz2023diffusionbased,
title={Diffusion-based Time Series Imputation and Forecasting with Structured State Space Models},
author={Juan Lopez Alcaraz and Nils Strodthoff},
journal={Transactions on Machine Learning Research},
issn={2835-8856},
year={2023},
note={}
}

@article{angelopoulos2021gentle,
  title={A gentle introduction to conformal prediction and distribution-free uncertainty quantification},
  author={Angelopoulos, Anastasios N and Bates, Stephen},
  journal={arXiv preprint arXiv:2107.07511},
  year={unpublished results}
}

@article{
bluecher2024decoupling,
title={Decoupling Pixel Flipping and Occlusion Strategy for Consistent {XAI} Benchmarks},
author={Stefan Bluecher and Johanna Vielhaben and Nils Strodthoff},
journal={Transactions on Machine Learning Research},
issn={2835-8856},
year={2024},
url={https://openreview.net/forum?id=bIiLXdtUVM},
note={}
}

@article{wagner2023explaining,
  title = {Explaining deep learning for ECG analysis: Building blocks for auditing and knowledge discovery},
  volume = {176},
  ISSN = {0010-4825},
  DOI = {10.1016/j.compbiomed.2024.108525},
  journal = {Computers in Biology and Medicine},
  publisher = {Elsevier BV},
  author = {Wagner,  Patrick and Mehari,  Temesgen and Haverkamp,  Wilhelm and Strodthoff,  Nils},
  year = {2024},
  month = jun,
  pages = {108525}
}

@article{shen2017deep,
  title={Deep learning in medical image analysis},
  author={Shen, Dinggang and Wu, Guorong and Suk, Heung-Il},
  journal={Annual review of biomedical engineering},
  volume={19},
  pages={221--248},
  year={2017},
  publisher={Annual Reviews}
}

@article{bepler2021learning,
  title={Learning the protein language: Evolution, structure, and function},
  author={Bepler, Tristan and Berger, Bonnie},
  journal={Cell systems},
  volume={12},
  number={6},
  pages={654--669},
  year={2021},
  publisher={Elsevier}
}

@article{Alcaraz:2023Synthetic,
  doi = {10.1016/j.compbiomed.2023.107115},
  year = {2023},
  month = jun,
  publisher = {Elsevier {BV}},
  pages = {107115},
  author = {Juan Miguel Lopez Alcaraz and Nils Strodthoff},
  title = {Diffusion-based conditional {ECG} generation with structured state space models},
  journal = {Computers in Biology and Medicine},
  eprint={2301.08227}
}

@inproceedings{Gu2021EfficientlyML,
  title={Efficiently Modeling Long Sequences with Structured State Spaces},
  author={Gu, Albert and Goel, Karan and R{\'e}, Christopher},
  booktitle={International Conference on Learning Representations},
  year={unpublished results},
        eprint={2111.00396},
      archivePrefix={arXiv},
      primaryClass={cs.LG}
}

@article{Wagner:2020PTBXL,
doi = {10.1038/s41597-020-0495-6},
year = {2020},
publisher = {Springer Science and Business Media {LLC}},
volume = {7},
number = {1},
pages = {154},
author = {Patrick Wagner and Nils Strodthoff and Ralf-Dieter Bousseljot and Dieter Kreiseler and Fatima I. Lunze and Wojciech Samek and Tobias Schaeffter},
title = {{PTB}-{XL},  a large publicly available electrocardiography dataset},
journal = {Scientific Data}
}

@article{Goldberger2020:physionet,
author = {Ary L. Goldberger  and Luis A. N. Amaral  and Leon Glass  and Jeffrey M. Hausdorff  and Plamen Ch. Ivanov  and Roger G. Mark  and Joseph E. Mietus  and George B. Moody  and Chung-Kang Peng  and H. Eugene Stanley },
title = {{PhysioBank, PhysioToolkit, and PhysioNet}},
journal = {Circulation},
volume = {101},
number = {23},
pages = {e215-e220},
year = {2000},
doi = {10.1161/01.CIR.101.23.e215}
}

@article{ismail2020inceptiontime,
  title={Inceptiontime: Finding alexnet for time series classification},
  author={Ismail Fawaz, Hassan and Lucas, Benjamin and Forestier, Germain and Pelletier, Charlotte and Schmidt, Daniel F and Weber, Jonathan and Webb, Geoffrey I and Idoumghar, Lhassane and Muller, Pierre-Alain and Petitjean, Fran{\c{c}}ois},
  journal={Data Mining and Knowledge Discovery},
  volume={34},
  number={6},
  pages={1936--1962},
  year={2020},
  publisher={Springer}
}

@misc{raykar2023tsshap,
      title={TsSHAP: Robust model agnostic feature-based explainability for time series forecasting}, 
      author={Vikas C. Raykar and Arindam Jati and Sumanta Mukherjee and Nupur Aggarwal and Kanthi Sarpatwar and Giridhar Ganapavarapu and Roman Vaculin},
      year={unpublished results},
      eprint={2303.12316},
      archivePrefix={arXiv},
      primaryClass={cs.LG}
}

@inproceedings{DBLP:conf/iclr/KongPHZC21,
  author    = {Zhifeng Kong and
               Wei Ping and
               Jiaji Huang and
               Kexin Zhao and
               Bryan Catanzaro},
  title     = {DiffWave: {A} Versatile Diffusion Model for Audio Synthesis},
  booktitle = {9th International Conference on Learning Representations, {ICLR} 2021},
  year      = {2021},
}

@article{Bluecher:2021PredDiff,
  doi = {10.1016/j.artint.2022.103774},
  year = {2022},
  publisher = {Elsevier {BV}},
  volume = {312},
  pages = {103774},
  author = {Stefan Bl\"{u}cher and Johanna Vielhaben and Nils Strodthoff},
  title = {{PredDiff}: Explanations and interactions from conditional expectations},
  journal = {Artificial Intelligence},
  eprint={2102.13519},
      archivePrefix={arXiv},
      primaryClass={cs.LG},
}

@InProceedings{pmlr-v70-shalit17a,
  title = 	 {Estimating individual treatment effect: generalization bounds and algorithms},
  author =       {Uri Shalit and Fredrik D. Johansson and David Sontag},
  booktitle = 	 {Proceedings of the 34th International Conference on Machine Learning},
  pages = 	 {3076--3085},
  year = 	 {2017},
  editor = 	 {Precup, Doina and Teh, Yee Whye},
  volume = 	 {70},
  series = 	 {Proceedings of Machine Learning Research},
  month = 	 {06--11 Aug},
  publisher =    {PMLR}
}

@article{goyal2020explaining,
      title={Explaining Classifiers with Causal Concept Effect (CaCE)}, 
      journal={arXiv preprint 1907.07165},
      author={Yash Goyal and Amir Feder and Uri Shalit and Been Kim},
      year={unpublished results},
      eprint={1907.07165},
      archivePrefix={arXiv},
      primaryClass={cs.LG}
}

@book{gillies2018causality,
  title={Causality, Probability, and Medicine},
  author={Gillies, D.},
  isbn={9781317564287},
  lccn={2020691788},
  year={2018},
  publisher={Taylor \& Francis}
}

@article{Wang_2019,
	doi = {10.1016/j.patrec.2018.02.010},
	year = 2019,
	month = {mar},
	publisher = {Elsevier {BV}},
	volume = {119},
	pages = {3--11},
	author = {Jindong Wang and Yiqiang Chen and Shuji Hao and Xiaohui Peng and Lisha Hu},
	title = {Deep learning for sensor-based activity recognition: A survey},
	journal = {Pattern Recognition Letters}
}

@article{rajkomar2018scalable,
  title={Scalable and accurate deep learning with electronic health records},
  author={Rajkomar, Alvin and Oren, Eyal and Chen, Kai and Dai, Andrew M and Hajaj, Nissan and Hardt, Michaela and Liu, Peter J and Liu, Xiaobing and Marcus, Jake and Sun, Mimi and others},
  journal={NPJ digital medicine},
  volume={1},
  number={1},
  pages={18},
  year={2018},
  publisher={Nature Publishing Group UK London}
}

@inproceedings{bashar2020tanogan,
  title={TAnoGAN: Time series anomaly detection with generative adversarial networks},
  author={Bashar, Md Abul and Nayak, Richi},
  booktitle={2020 IEEE Symposium Series on Computational Intelligence (SSCI)},
  pages={1778--1785},
  year={2020},
  organization={IEEE}
}

@inproceedings{hssayeni2022imbalanced,
  title={Imbalanced time-series data regression using conditional generative adversarial networks},
  author={Hssayeni, Murtadha D},
  booktitle={International Conference on Machine Learning and Applications},
  year={2022}
}

@article{tashiro2021csdi,
  title={CSDI: Conditional score-based diffusion models for probabilistic time series imputation},
  author={Tashiro, Yusuke and Song, Jiaming and Song, Yang and Ermon, Stefano},
  journal={Advances in Neural Information Processing Systems},
  volume={34},
  pages={24804--24816},
  year={2021}
}

@article{borisov2005analysis,
  title={Analysis of EEG structural synchrony in adolescents with schizophrenic disorders},
  author={Borisov, SV and Kaplan, A Ya and Gorbachevskaya, NL and Kozlova, IA},
  journal={Human Physiology},
  volume={31},
  pages={255--261},
  year={2005},
  publisher={Springer}
}

@article{2022_resting, title={Bayesian Optimization of Machine Learning Classification of Resting-State EEG Microstates in Schizophrenia: A Proof-of-Concept Preliminary Study Based on Secondary Analysis}, volume={12}, ISSN={2076-3425}, 
DOI={10.3390/brainsci12111497}, number={11}, journal={Brain Sciences}, publisher={MDPI AG}, author={Keihani, Ahmadreza and Sajadi, Seyed Saman and Hasani, Mahsa and Ferrarelli, Fabio}, year={2022}, month={Nov}, pages={1497} }

@ARTICLE{10.3389/fpsyt.2021.761203,
  
AUTHOR={Sun, Qiaoling and Zhou, Jiansong and Guo, Huijuan and Gou, Ningzhi and Lin, Ruoheng and Huang, Ying and Guo, Weilong and Wang, Xiaoping},   
	 
TITLE={EEG Microstates and Its Relationship With Clinical Symptoms in Patients With Schizophrenia},      
	
JOURNAL={Frontiers in Psychiatry},      
	
VOLUME={12},           
	
YEAR={2021},      
	  	
DOI={10.3389/fpsyt.2021.761203},      
	
ISSN={1664-0640}
}

@article{ismail2020benchmarking,
  title={Benchmarking deep learning interpretability in time series predictions},
  author={Ismail, Aya Abdelsalam and Gunady, Mohamed and Corrada Bravo, Hector and Feizi, Soheil},
  journal={Advances in neural information processing systems},
  volume={33},
  pages={6441--6452},
  year={2020}
}

@article{zhao2023interpretation,
  title={Interpretation of Time-Series Deep Models: A Survey},
  author={Zhao, Ziqi and Shi, Yucheng and Wu, Shushan and Yang, Fan and Song, Wenzhan and Liu, Ninghao},
  journal={arXiv preprint arXiv:2305.14582},
  year={unpublished results}
}

@article{rojat2021explainable,
  title={Explainable artificial intelligence (xai) on timeseries data: A survey},
  author={Rojat, Thomas and Puget, Rapha{\"e}l and Filliat, David and Del Ser, Javier and Gelin, Rodolphe and D{\'\i}az-Rodr{\'\i}guez, Natalia},
  journal={arXiv preprint arXiv:2104.00950},
  year={unpublished results}
}

@inproceedings{crabbe2021explaining,
  title={Explaining time series predictions with dynamic masks},
  author={Crabb{\'e}, Jonathan and Van Der Schaar, Mihaela},
  booktitle={International Conference on Machine Learning},
  pages={2166--2177},
  year={2021},
  organization={PMLR}
}

@inproceedings{Schoelkopf2012,
author = {Sch\"{o}lkopf, Bernhard and Janzing, Dominik and Peters, Jonas and Sgouritsa, Eleni and Zhang, Kun and Mooij, Joris},
title = {On Causal and Anticausal Learning},
year = {2012},
isbn = {9781450312851},
publisher = {Omnipress},
address = {Madison, WI, USA},
booktitle = {Proceedings of the 29th International Coference on International Conference on Machine Learning},
pages = {459–466},
numpages = {8},
location = {Edinburgh, Scotland},
series = {ICML'12}
}

@article{lundberg2017unified,
	title={A unified approach to interpreting model predictions},
	author={Lundberg, Scott M and Lee, Su-In},
	journal={Advances in Neural Information Processing Systems},
	year={2017},
	volume={30},
}

@Article{Montavon2018,
  author    = {Montavon, Gr{\'e}goire and Samek, Wojciech and M{\"u}ller, Klaus-Robert},
  journal   = {Digital Signal Processing},
  title     = {Methods for interpreting and understanding deep neural networks},
  year      = {2018},
  pages     = {1--15},
  volume    = {73},
  publisher = {Elsevier},
}

@Article{covert2021explaining,
  author  = {Covert, Ian and Lundberg, Scott and Lee, Su-In},
  journal = {Journal of Machine Learning Research},
  title   = {Explaining by removing: A unified framework for model explanation},
  year    = {2021},
  number  = {209},
  pages   = {1--90},
  volume  = {22},
}

@article{thygesen2018fourth,
  title={Fourth universal definition of myocardial infarction (2018)},
  author={Thygesen, Kristian and Alpert, Joseph S and Jaffe, Allan S and Chaitman, Bernard R and Bax, Jeroen J and Morrow, David A and White, Harvey D and Executive Group on behalf of the Joint European Society of Cardiology (ESC)/American College of Cardiology (ACC)/American Heart Association (AHA)/World Heart Federation (WHF) Task Force for the Universal Definition of Myocardial Infarction},
  journal={Circulation},
  volume={138},
  number={20},
  pages={e618--e651},
  year={2018},
  publisher={Am Heart Assoc}
}

@article{esteva2019guide,
  title={A guide to deep learning in healthcare},
  author={Esteva, Andre and Robicquet, Alexandre and Ramsundar, Bharath and Kuleshov, Volodymyr and DePristo, Mark and Chou, Katherine and Cui, Claire and Corrado, Greg and Thrun, Sebastian and Dean, Jeff},
  journal={Nature medicine},
  volume={25},
  number={1},
  pages={24--29},
  year={2019},
  publisher={Nature Publishing Group US New York}
}

@article{miotto2018deep,
  title={Deep learning for healthcare: review, opportunities and challenges},
  author={Miotto, Riccardo and Wang, Fei and Wang, Shuang and Jiang, Xiaoqian and Dudley, Joel T},
  journal={Briefings in bioinformatics},
  volume={19},
  number={6},
  pages={1236--1246},
  year={2018},
  publisher={Oxford University Press}
}

@misc{Minixhofer_2021, title={Predict droughts using weather \& soil data}, url={https://www.kaggle.com/datasets/cdminix/us-drought-meteorological-data}, journal={Kaggle}, author={Minixhofer, Christoph}, year={2021}, month={Mar}}

@article{GramfortEtAl2013a,
  title = {{{MEG}} and {{EEG}} Data Analysis with {{MNE}}-{{Python}}},
  author = {Gramfort, Alexandre and Luessi, Martin and Larson, Eric and Engemann, Denis A. and Strohmeier, Daniel and Brodbeck, Christian and Goj, Roman and Jas, Mainak and Brooks, Teon and Parkkonen, Lauri and H{\"a}m{\"a}l{\"a}inen, Matti S.},
  year = {2013},
  volume = {7},
  pages = {1--13},
  doi = {10.3389/fnins.2013.00267},
  journal = {Frontiers in Neuroscience},
  number = {267}
}

@misc{githubGitHubFredericvWeegmicrostates,
  author = {Frederic von Wegner},
  title = {GitHub - EEG microstate analysis -github.com},
  url = {https://github.com/Frederic-vW/eeg\_microstates/tree/master},
  year = {2017},
  note = {[Accessed 28-04-2024]},
}

@inproceedings{ates2021counterfactual,
  title={Counterfactual explanations for multivariate time series},
  author={Ates, Emre and Aksar, Burak and Leung, Vitus J and Coskun, Ayse K},
  booktitle={2021 international conference on applied artificial intelligence (ICAPAI)},
  pages={1--8},
  year={2021},
  organization={IEEE}
}

@inproceedings{delaney2021instance,
  title={Instance-based counterfactual explanations for time series classification},
  author={Delaney, Eoin and Greene, Derek and Keane, Mark T},
  booktitle={International conference on case-based reasoning},
  pages={32--47},
  year={2021},
  organization={Springer}
}

@inproceedings{wang2021learning,
  title={Learning time series counterfactuals via latent space representations},
  author={Wang, Zhendong and Samsten, Isak and Mochaourab, Rami and Papapetrou, Panagiotis},
  booktitle={Discovery Science: 24th International Conference, DS 2021, Halifax, NS, Canada, October 11--13, 2021, Proceedings 24},
  pages={369--384},
  year={2021},
  organization={Springer}
}

@inproceedings{li2022motif,
  title={Motif-guided time series counterfactual explanations},
  author={Li, Peiyu and Boubrahimi, Souka{\"\i}na Filali and Hamdi, Shah Muhammad},
  booktitle={International Conference on Pattern Recognition},
  pages={203--215},
  year={2022},
  organization={Springer}
}

@article{pascual1995segmentation,
  title={Segmentation of brain electrical activity into microstates: model estimation and validation},
  author={Pascual-Marqui, Roberto D and Michel, Christoph M and Lehmann, Dietrich},
  journal={IEEE Transactions on Biomedical Engineering},
  volume={42},
  number={7},
  pages={658--665},
  year={1995},
  publisher={IEEE}
}

@article{10.1093/ehjdh/ztae039,
    author = {Strodthoff, Nils and Lopez Alcaraz, Juan Miguel and Haverkamp, Wilhelm},
    title = "{Prospects for AI-Enhanced ECG as a Unified Screening Tool for Cardiac and Non-Cardiac Conditions – An Explorative Study in Emergency Care}",
    journal = {European Heart Journal - Digital Health},
    pages = {ztae039},
    year = {2024},
    month = {05},
    issn = {2634-3916},
    doi = {10.1093/ehjdh/ztae039},
    eprint = {https://academic.oup.com/ehjdh/advance-article-pdf/doi/10.1093/ehjdh/ztae039/57553846/ztae039.pdf},
}

@article{KIKUCHI2007163,
title = {Native EEG and treatment effects in neuroleptic-naïve schizophrenic patients: Time and frequency domain approaches},
journal = {Schizophrenia Research},
volume = {97},
number = {1},
pages = {163-172},
year = {2007},
issn = {0920-9964},
doi = {https://doi.org/10.1016/j.schres.2007.07.012},
author = {Mitsuru Kikuchi and Thomas Koenig and Yuji Wada and Masato Higashima and Yoshifumi Koshino and Werner Strik and Thomas Dierks}
}

@article{koenig1999deviant,
  title={A deviant EEG brain microstate in acute, neuroleptic-naive schizophrenics at rest},
  author={Koenig, Thomas and Lehmann, Dietrich and Merlo, Marco CG and Kochi, Kieko and Hell, Daniel and Koukkou, Martha},
  journal={European archives of psychiatry and clinical neuroscience},
  volume={249},
  pages={205--211},
  year={1999},
  publisher={Springer}
}

@article{NISHIDA20131106,
  title = {EEG microstates associated with salience and frontoparietal networks in frontotemporal dementia, schizophrenia and Alzheimer’s disease},
  journal = {Clinical Neurophysiology},
  volume = {124},
  number = {6},
  pages = {1106-1114},
  year = {2013},
  issn = {1388-2457},
  doi = {https://doi.org/10.1016/j.clinph.2013.01.005},
  author = {Keiichiro Nishida and Yosuke Morishima and Masafumi Yoshimura and Toshiaki Isotani and Satoshi Irisawa and Kay Jann and Thomas Dierks and Werner Strik and Toshihiko Kinoshita and Thomas Koenig}
}

@article{dressler1947high,
  title={High T waves in the earliest stage of myocardial infarction},
  author={Dressler, William and Hugo, Roesler},
  journal={American heart journal},
  volume={34},
  number={5},
  pages={627--645},
  year={1947},
  publisher={Elsevier}
}

@article{grossman1969serial,
  title={Serial P wave changes in acute myocardial infarction},
  author={Grossman, James I and Delman, Abner J},
  journal={American Heart Journal},
  volume={77},
  number={3},
  pages={336--341},
  year={1969},
  publisher={Elsevier}
}

@article{cancelliere2007drought,
  title={Drought forecasting using the standardized precipitation index},
  author={Cancelliere, Antonino and Mauro, G Di and Bonaccorso, Brunella and Rossi, G},
  journal={Water resources management},
  volume={21},
  pages={801--819},
  year={2007},
  publisher={Springer}
}

@article{anshuka2019drought,
  title={Drought forecasting through statistical models using standardised precipitation index: a systematic review and meta-regression analysis},
  author={Anshuka, Anshuka and van Ogtrop, Floris F and Willem Vervoort, R},
  journal={Natural Hazards},
  volume={97},
  pages={955--977},
  year={2019},
  publisher={Springer}
}

@article{karim2017lstm,
  title={LSTM fully convolutional networks for time series classification},
  author={Karim, Fazle and Majumdar, Somshubra and Darabi, Houshang and Chen, Shun},
  journal={IEEE access},
  volume={6},
  pages={1662--1669},
  year={2017},
  publisher={IEEE}
}

@article{russwurm2020self,
  title={Self-attention for raw optical satellite time series classification},
  author={Ru{\ss}wurm, Marc and K{\"o}rner, Marco},
  journal={ISPRS journal of photogrammetry and remote sensing},
  volume={169},
  pages={421--435},
  year={2020},
  publisher={Elsevier}
}

@inproceedings{rakthanmanon2013fast,
  title={Fast shapelets: A scalable algorithm for discovering time series shapelets},
  author={Rakthanmanon, Thanawin and Keogh, Eamonn},
  booktitle={proceedings of the 2013 SIAM International Conference on Data Mining},
  pages={668--676},
  year={2013},
  organization={SIAM}
}

@article{fulcher2017hctsa,
  title={hctsa: A computational framework for automated time-series phenotyping using massive feature extraction},
  author={Fulcher, Ben D and Jones, Nick S},
  journal={Cell systems},
  volume={5},
  number={5},
  pages={527--531},
  year={2017},
  publisher={Elsevier}
}

@article{deng2013time,
  title={A time series forest for classification and feature extraction},
  author={Deng, Houtao and Runger, George and Tuv, Eugene and Vladimir, Martyanov},
  journal={Information Sciences},
  volume={239},
  pages={142--153},
  year={2013},
  publisher={Elsevier}
}

@article{hills2014classification,
  title={Classification of time series by shapelet transformation},
  author={Hills, Jon and Lines, Jason and Baranauskas, Edgaras and Mapp, James and Bagnall, Anthony},
  journal={Data mining and knowledge discovery},
  volume={28},
  pages={851--881},
  year={2014},
  publisher={Springer}
}

@article{schafer2015boss,
  title={The BOSS is concerned with time series classification in the presence of noise},
  author={Sch{\"a}fer, Patrick},
  journal={Data Mining and Knowledge Discovery},
  volume={29},
  pages={1505--1530},
  year={2015},
  publisher={Springer}
}

@article{alcaraz2024mds,
  title={MDS-ED: Multimodal Decision Support in the Emergency Department--a Benchmark Dataset for Diagnoses and Deterioration Prediction in Emergency Medicine},
  author={Alcaraz, Juan Miguel Lopez and Strodthoff, Nils},
  journal={arXiv preprint arXiv:2407.17856},
  year={unpublished results}
}

@inproceedings{ma2024tsfeatlime,
  title={TSFeatLIME: An Online User Study in Enhancing Explainability in Univariate Time Series Forecasting},
  author={Ma, Hongnan and McAreavey, Kevin and Liu, Weiru},
  booktitle={2024 IEEE 36th International Conference on Tools with Artificial Intelligence (ICTAI)},
  pages={578--585},
  year={2024},
  organization={IEEE}
}

\appendix
\section{Appendix}

\subsection{Computational complexity}\label{app:computation}

\begin{table}[!ht]
    \caption{Computational complexity}
    \label{tab:computational-complexity}
    \centering
    \begin{tabular}{lcc}
        \toprule
        \textbf{Description} & \textbf{Train[h]} & \textbf{Generate[s]} \\
        \midrule
        Drought & 18.5 & 21 \\ 
        PTB-XL & 18.9 & 21 \\
        Schizophrenia & 19.2  & 21 \\
        \bottomrule 
    \end{tabular}
\end{table}

Tab.~\ref{tab:computational-complexity} contains the details of the computational complexity of the proposed approach. We present the training in hours and sampling in seconds of each dataset separately as certain attributes differ from each other, such as the sample length, and the number of channels of each time series, similarly, one has to consider the computational power in use, in this case, the model training was executed on separate NVIDIA L40 GPUs, each with 48GB VRAM and around 18,176 CUDA cores, supported by 16GB RAM and 16-CPU cores. The results of the generation column represent the time for the imputation of a single concept for the class.

\subsection{Datasets}\label{app: datasets}

\begin{table}[!ht]
    \caption{Datasets details}
    \label{tab:dataset-details}
    \centering
    \begin{tabular}{lccc}
        \toprule
        \textbf{Description}  & \textbf{Drought} & \textbf{PTB-XL} & \textbf{Schizophrenia}  \\
        \midrule
        Train size   & 300,000  & 9,754  & 1,980  \\ 
        Validation size & 165,638  &  1,226 & 270 \\ 
        Test size & 169,450 & 1,232  & 270 \\ 
        Classifier batch & 32 & 32  &  32  \\
        Imputer batch & 6   &  6   & 6  \\
        Sample length & 180  & 248 & 248  \\
        Sample features&  18 & 12  & 16  \\ 
        Classes & 2 &  2   & 2 \\ 
        Concepts & 4 &  6   &  4   \\
        \bottomrule
    \end{tabular}
\end{table}

Tab.~\ref{tab:dataset-details} provides details for the three considered datasets: Drought, PTB-XL, and Schizophrenia. It includes information on the size of the training, validation, and test sets, batch sizes for both classifier and imputer models, sample length, number of sample features, classes, and concepts present in each dataset. To avoid data leakage, the drought dataset is split by the provided time horizons from past to present into train, val, and test, whereas the PTB-XL and schizophrenia datasets are split patient-wise. For the Drought dataset, concepts were generated using k-means with elbow-method leveraging the implementation from sci-kit learn. PTB-XL utilized well-defined concepts from existing literature \cite{wagner2023explaining}. Meanwhile, for the schizophrenia dataset, we employed a micro-states open-source software that is internally based on k-means, where based on the elbow method we select 4 microstate concepts.

\subsection{Models}\label{app: models}

\begin{table}[!ht]
    \caption{S4 hyperparameters}
    \label{tab:s4-hyperparameters}
    \centering
    \begin{tabular}{ll}
        \toprule
        \textbf{Hyperparameter} & \textbf{Value}  \\
        \midrule
         Block of layers & 4 \\
         s4 model copies & 512 \\
         s4 state size & 8 \\ 
         Optimizer & Adam \\ 
         Learning rate & 0.001 \\ 
         Weight decay & 0.001 \\ 
         learning rate schedule & constant \\ 
         Batch size & 64 \\ 
         Epochs & 20 \\ 
        \bottomrule
    \end{tabular}
\end{table}

Tab.~\ref{tab:s4-hyperparameters} outlines the hyperparameters employed in the S4 model. The architecture consists of four blocks of layers, with each block containing 512 copies of the S4 model. The state size within the S4 model is set to 8. For optimization, the Adam optimizer is utilized with a learning rate and weight decay both set to 0.001. The learning rate schedule is maintained constant throughout training. A batch size of 64 samples is used for each training iteration, spanning a total of 20 epochs. The training objective is to minimize the binary cross-entropy loss. During training, we apply a model selection strategy on the best performance (AUROC) on the validation set.

\begin{table}[!ht]
    \caption{Diffusion model hyperparameters}
    \label{tab:diffusion-model-hyperparameters}
    \centering
    \begin{tabular}{ll}
        \toprule
        \textbf{Hyperparameter} & \textbf{Value} \\
        \midrule
        Residual layers & 36 \\ 
        Residual channels & 256 \\
        Skip channels & 256 \\ 
        Diffusion embedding dim. 1 & 128 \\ 
        Diffusion embedding dim. 2 & 512 \\ 
        Diffusion embedding dim. 3 & 512 \\ 
        Schedule  & Linear \\ 
        Diffusion steps $T$ & 200 \\ 
        $B_0$ & 0.0001 \\ 
        $B_1$ & 0.02 \\ 
        Optimizer & Adam \\ 
        Loss function & MSE \\
        Learning rate & 0.0002 \\
        S4 state $N$ dimensions & 64 \\ 
        S4 bidirectional & Yes  \\ 
        S4 layer normalization & Yes \\ 
        S4 Drop-out & 0.0  \\ 
        S4 Maximum length  & as required \\
        \bottomrule
    \end{tabular}
\end{table}

Tab.~\ref{tab:diffusion-model-hyperparameters} present the hyperparameters and training approach for the CausalConceptTS model. Built upon \textit{DiffWave} \cite{DBLP:conf/iclr/KongPHZC21}, and previously presented as \textit{SSSD} \cite{lopezalcaraz2023diffusionbased} our model consists of 36 residual layers with 256 channels. It integrates a three-layer diffusion embedding (128, 256, and 256 dimensions) with swish activations, followed by convolutional layers. Our diffusion spans 200 time steps, using a linear schedule from 0.0001 to 0.02 for beta. We optimize with Adam (LR: 0.0002). Based on previous works \cite{lopezalcaraz2023diffusionbased} we trained each model over 50,000 iterations with model selection on lower MSE loss every 1,000 iterations. For the S4 model, we utilize a bidirectional layer with layer normalization, no dropout, and internal state dimension $N=64$. This S4 layer captures bidirectional time dependencies. We maintain layer normalization and an internal state of $N=64$, consistent with prior work \cite{Gu2021EfficientlyML}. 

In this work, we trained class-specific imputer models, one for each condition under consideration. An obvious alternative might seem to be to train a class-conditional imputer model. However, this requires to specify a procedure to adjust the importance of the class-conditional input within the framework of classifier-free guidance. While we observed minimal effects on dropout rates during training, increasing the alpha parameter during sampling improved imputation and causal effects but resulted in unrealistic time series, like excessively large R peaks in ECG data. As a result, we decided to base our experiments on class-specific imputation models.

\subsection{Comparing associational and causal attributions: single-concept toy example}\label{app:toy}

\heading{Setup} Consider a simple example where the signal is represented by a sine wave with amplitude $a\in[0,2]$. We assume that $a\geq 1$ corresponds to the positive and $a<1$ to the negative class. More specifically, we assume that the classifiers output probability is given by $f(a)=\frac{a}{2}$. Furthermore, the data generative process is parametrized via $$p(a)=\begin{cases}
p & \text{if } a \in [0,1] \\
1-p & \text{if } a \in [1,2]\,.
\end{cases}$$ Here, the parameter $p$ allows to induce a class imbalance between the positive and the negative class.

\heading{Causal Attribution}
We can now work out the causal attribution as follows
\begin{align}
ITE&=\int_1^2 \log_2(\frac{a}{2})\text{d}a-\int_1^2 \log_2(\frac{a}{2})\text{d}a\nonumber\\
&=1-\frac{1}{\log 2}+1+\frac{1}{\log 2}=2\,.
\end{align}
It turns out to be positive irrespective of the value $a$ in the input sample under consideration.

\heading{Associational Attribution}
Similarly, we can work out the associational attribution for a sample with parameter value $a$:
\begin{align}
IAA&=\log_2 f(a)-\int_0^2 p(a) f(a)\text{d}a\nonumber\\
&=\log_2 \frac{a}{2} -p \left(-1-\frac{1}{\log 2}\right)-(1-p)\left(1-\frac{1}{\log 2}\right)\,.
\end{align}
Let us now consider the attribution in the case of a sample from the positive class, i.e., $a\geq 1$. The first terms is minimal for $a=1$ and it is straightforward to work out that the attribution changes sign within the positive class if $p>1-\frac{1}{2\log 2}\approx 0.279$, i.e., in the case where the data distribution shows a sufficient imbalance towards the positive class.

\heading{Conclusion}
We argue that this sign change in the association attribution is an unnatural behaviour for the classifier defined above, which is not observed in the case of the causal attribution. We fully acknowledge that this is a very simple example and that more complicated examples should involve correlations between features/concepts and consider imperfect classifiers that at least partially leverage such correlations.

\end{document}